\documentclass[runningheads]{llncs}
\usepackage{graphicx}
\usepackage{amsmath,amssymb} % define this before the line numbering.
\usepackage{color}
\usepackage{booktabs}
\usepackage{multirow}
\usepackage{cite}

\begin{document}
\title{Pairwise Relational Networks for Face Recognition}
% Replace with your title

\titlerunning{Pairwise Relational Networks for Face Recognition}
% Replace with a meaningful short version of your title
%
\author{Bong-Nam Kang\inst{1}\orcidID{0000-0002-6818-7532} \and
Yonghyun Kim\inst{2}\orcidID{0000-0003-0038-7850} \and
Daijin Kim\inst{1,2}\orcidID{0000-0002-8046-8521}}
%
%Please write out author names in full in the paper, i.e. full given and family names.
%If any authors have names that can be parsed into FirstName LastName in multiple ways, please include the correct parsing, in a comment to the volume editors:
%\index{Lastnames, Firstnames}
%(Do not uncomment it, because you may introduce extra index items if you do that, we will use scripts for introducing index entries...)
\authorrunning{B-.N. Kang et al.}
% Replace with shorter version of the author list. If there are more authors than fits a line, please use A. Author et al.
%
\institute{Department of Creative IT Engineering, POSTECH, Korea\and
	Department of Computer Science and Engineering, POSTECH, Korea\\
	\email{\{bnkang,gkyh0805,dkim\}@postech.ac.kr}}

\maketitle              % typeset the header of the contribution
\begin{abstract}
Existing face recognition using deep neural networks is difficult to know what kind of features are used to discriminate the identities of face images clearly. To investigate the effective features for face recognition, we propose a novel face recognition method, called a pairwise relational network (PRN), that obtains local appearance patches around landmark points on the feature map, and captures the pairwise relation between a pair of local appearance patches. The PRN is trained to capture unique and discriminative pairwise relations among different identities. Because the existence and meaning of pairwise relations should be identity dependent, we add a face identity state feature, which obtains from the long short-term memory (LSTM) units network with the sequential local appearance patches on the feature maps, to the PRN. To further improve accuracy of face recognition, we combined the global appearance representation with the pairwise relational feature. Experimental results on the LFW show that the PRN using only pairwise relations achieved $99.65\%$ accuracy and the PRN using both pairwise relations and face identity state feature achieved $99.76\%$ accuracy. On the YTF, both the PRN using only pairwise relations and the PRN using pairwise relations and the face identity state feature achieved the \textit{state-of-the-art} ($95.7\%$ and $96.3\%$). The PRN also achieved comparable results to the \textit{state-of-the-art} for both face verification and face identification tasks on the IJB-A, and the \textit{state-of-the-art} on the IJB-B.

\end{abstract}
\section{Introduction}
Convolutional neural networks (CNNs) have achieved great success on computer vision community by improving the \textit{state-of-the-art} in almost all of applications, especially in classification problems including object \cite{DenseNet,ResNet,DeepRectifier,ImageNet,VGGNet,VeryDeepSmall,Googlenet} scene\cite{Scene1,Scene2}, and so on. The key to success of CNNs is the availability of large scale of training data and the end-to-end learning framework. The most commonly used CNNs perform feature learning and prediction of label information by mapping the input raw data to deep embedded features which are commonly the output of the last fully connected (FC) layer, and then predict labels using these deep embedded features. These approaches use the deep embedded features holistically for their applications, without knowing what part of the features is used and what it is meaning.

Face recognition in unconstrained environments is a very challenging problem in computer vision. Faces of the same identity can look very different when presented in different illuminations, facial poses, facial expressions, and occlusions. Such variations within the same identity could overwhelm the variations due to identity differences and make face recognition challenging. To solve these problems, many deep learning-based approaches have been proposed and achieved high accuracies of face recognition such as DeepFace \cite{DeepFace}, DeepID series \cite{DeepID,DeepID2,DeepID2+,DeepID3}, FaceNet \cite{FaceNet}, PIMNet \cite{PIMNet_CVPRW2017}, SphereFace \cite{SphereFace}, and ArcFace \cite{ArcFace}.

In face recognition tasks in unconstrained environments, the deeply learned and embedded features need to be not only separable but also discriminative. However, these features are learned implicitly for separable and distinct representations to classify among different identities without what part of the features is used, what part of the feature is meaningful, and what part of the features is separable and discriminative. Therefore, it is difficult to know what kind of features are used to discriminate the identities of face images clearly.

\begin{figure}[t]
	\centering
	\includegraphics[scale=0.25]{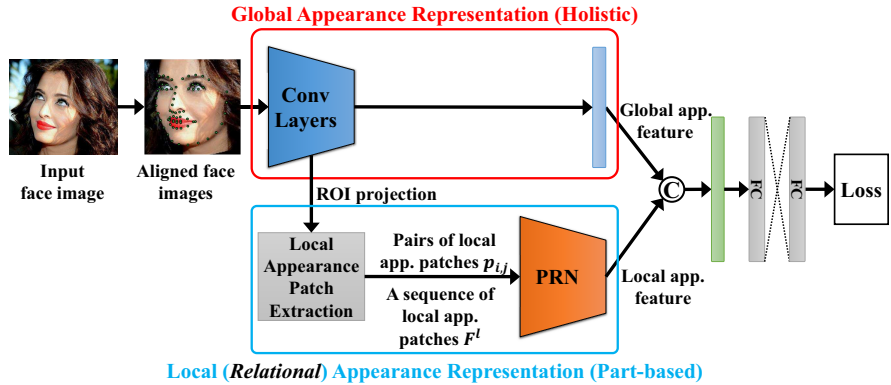}
	\caption{Overview of the proposed face recognition method}\label{fig:fig01_overview}
\end{figure}

To overcome this limitation, we propose a novel face recognition method, called a pairwise relational network (PRN) to capture unique relations within same identity and discriminative relations among different identities. To capture relations, the PRN takes local appearance patches as input by ROI projection around landmark points on the feature map in a backbone CNN network. With these local appearance patches, the PRN is trained to capture unique pairwise relations between pairs of local appearance patches to determine facial part-relational structures and properties in face images. Because the existence and meaning of pairwise relations should be identity dependent, the PRN could condition its processing on a facial identity state feature. The facial identity state feature is learned from the long short-term memory (LSTM) units network with the sequential local appearance patches on the feature maps. To further improve accuracy of face recognition, we combined the global appearance representation with the local appearance representation (the relation features) (\figurename~\ref{fig:fig01_overview}). More details of the proposed face recognition method are given in Section \ref{sec:proposed_methods}.

The main contributions of this paper can be summarized as follows:
\begin{itemize}
	\item We propose a novel face recognition method using the pairwise relational network (PRN) which captures the unique and discriminative pairwise relations of local appearance patches on the feature maps to classify face images among different identities.
	\item We show that the proposed PRN is very useful to enhance the accuracy of both face verification and face identification.
	\item We present extensive experiments on the public available datasets such as Labeled Faces in the Wild (LFW), YouTube Faces (YTF), IARPA Janus Benchmark-A (IJB-A), and IARPA Janus Benchmark-B (IJB-B).
\end{itemize}

The rest of this paper is as follows: in Section \ref{sec:proposed_methods} we describe the proposed face recognition method including the base CNN architecture, face alignment, pairwise relational network, facial identity state feature, loss function used for training the proposed method, respectively; in Sections \ref{sec:expt} we present experimental results of the proposed method in comparison with the \textit{state-of-the-art} on the public benchmark dataset and discussion; in Section \ref{sec:conclusion} we draw a conclusion.

\section{Proposed Methods}\label{sec:proposed_methods}
In this section, we describe our methods in detail including the base CNN model as backbone network for the global appearance representation, the face alignment method, the pairwise relational network, the pairwise relational network with face identity states, and the loss functions.

\subsection{Base Convolutional Neural Network}\label{sec:base_cnn}
We first describe the base CNN model. It is the backbone neural network to represent the global appearance representation and extract the local appearance patches to capture the relations (\figurename{~\ref{fig:fig01_overview}}).
The base CNN model consists of several 3-layer residual bottleneck blocks similar to the ResNet-101 \cite{ResNet}. The ResNet-101 has one convolution layer, one max pooling layer, $30$ 3-layer residual bottleneck blocks, one global average pooling (GAP) layer, one FC layer, and \textit{softmax} loss layer. The ResNet-101 accepts a image with $224\times224$ resolution as input, and has $7\times7$ convolution filters with a stride of 2 in the first layer. In contrast, our base CNN model accepts a face image with $140\times140$ resolution as input, and has $5\times5$ convolution filters with a stride of 1 in the first layer (\textit{conv1} in \tablename{~\ref{tab:tab_baseline_cnn}}). Because of different input resolution, size of kernel filters, and stride, the output size in each intermediate layer is also different from the original ResNet-101. In the last layer, we use the GAP with $9\times9$ filter in each channel and the FC layer. The outputs of FC layer are fed into the \textit{softmax} loss layer. More details of the base CNN architecture are given in \tablename{~\ref{tab:tab_baseline_cnn}}.

To represent the global appearance representation $\boldsymbol{f}^{g}$, we use the $1\times1\times2048$ feature which is the output of the GAP in the base CNN (\tablename{~\ref{tab:tab_baseline_cnn}}).

To represent the local appearance representation, we extract the local appearance patches $\boldsymbol{f}^{l}$ on the $9\times9\times2048$ feature maps (\textit{conv5\_3}) in the base CNN (\tablename{~\ref{tab:tab_baseline_cnn}}) by ROI projection with facial landmark points. These $\boldsymbol{f}^{l}$ are used to capture and model pairwise relations between them. More details of the local appearance patches and relations are described in Section \ref{sec:PRN}.

\begin{table}[t]
	\centering
	\caption{Base convolutional neural network. The base CNN is similar to ResNet-101, but the dimensionality of input, the size of convolution filters, and the size of each output feature map are different from the original ResNet-101}
	\label{tab:tab_baseline_cnn}
	\begin{tabular}{@{}c|c|c@{}}
		\toprule
		\textbf{Layer name}    & \textbf{Output size} & \textbf{101-layer} \\ \hline
		\textit{conv1}         & $140\times140$       & $5\times5$, $64$         \\ \hline
		\multirow{2}{*}{\textit{conv2\_x}} & \multirow{2}{*}{$70\times70$} & $3\times3$ max pool, stride $2$       \\ \cline{3-3}
		&                      & $\left[\begin{array}{c} 1\times1, 64 \\ 3\times3, 64 \\ 1\times1, 256 \end{array} \right] \times3$     \\ \hline
		\textit{conv3\_x}      & $35\times35$         & $\left[\begin{array}{c} 1\times1, 128 \\ 3\times3, 128 \\ 1\times1, 512 \end{array} \right] \times4$   \\ \hline
		\textit{conv4\_x}      & $18\times18$         & $\left[\begin{array}{c} 1\times1, 256 \\ 3\times3, 256 \\ 1\times1, 1024 \end{array} \right] \times23$  \\ \hline
		\textit{conv5\_x}      & $9\times9$           & $\left[\begin{array}{c} 1\times1, 512 \\ 3\times3, 512 \\ 1\times1, 2048 \end{array} \right] \times3$  \\ \hline
		& $1\times1$           & global average pool, 8630-d fc, \textit{softmax}         \\ \bottomrule
	\end{tabular}
\end{table}

\subsection{Face Alignment}\label{sec:face_alignment}
In the base CNN model, the input layer accepts the RGB values of the face image pixels. We employ a face alignment method to align a face image into the canonical face image, then we adopt this aligned face image as input of our base CNN model.

\begin{figure}[t]
	\centering
	\includegraphics[scale=0.40]{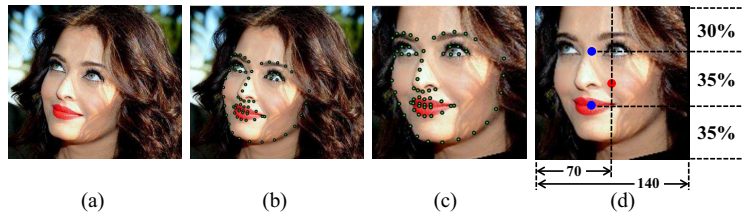}
	\caption{A face alignment. The original image is shown in (a); (b) shows the detected $68$ landmark points; (c) shows the aligned $68$ landmark points in the aligned image plane; and (d) is the final aligned face image, where the red circle was used to center the face image along $x$-axis, and the blue circles denote the two points used for face cropping}
	\label{fig:fig_alignment}
\end{figure}

The alignment procedures are as follows: 1) Use the DAN implementation of Kowalsky \textit{et al.} by using multi-stage neural network \cite{DAN_CVPRW2017} to detect $68$ facial landmark points (\figurename{~\ref{fig:fig_alignment}b}); 2) rotate the face in the image plane to make it upright based on the eye positions (\figurename{~\ref{fig:fig_alignment}c}); 3) find a central point on the face by taking the mid-point between the leftmost and rightmost landmark points (the red point in \figurename{~\ref{fig:fig_alignment}d}); 4) the center points of the eye and mouth (blue points in \figurename{~\ref{fig:fig_alignment}d}) are found by averaging all the landmark points in the eye and mouth regions, respectively; 5) center the faces in the $x$-axis, based on the center point (red point); 6) fix the position along the $y$-axis by placing the eye center point at $30\%$ from the top of the image and the mouth center point at $35\%$ from the bottom of the image, respectively; 7) resize the image to a resolution of $140\times140$.
Each pixel which value is in a range of $[0, 255]$ in the RGB color space is normalized by dividing $255$ to be in a range of $[0, 1]$.

\subsection{Pairwise Relational Network}\label{sec:PRN}
The pairwise relational network (PRN) is a neural network and takes a set of local appearance patches on the feature maps as input and output a single feature vector as its relational feature for the face recognition task. The PRN captures unique pairwise relations between pairs of local appearance patches within the same identity and discriminative pairwise relations among different identities.
In other words, the PRN captures the core common properties of faces within the same identity, while captures the discriminative properties of faces among different identities. Therefore, the PRN aims to determine pairwise-relational structures from pairs of local appearance patches in face images.
The relation feature $\boldsymbol{r}_{i,j}$ represents a latent relation of a pair of two local appearance patches, and can be written as follows:
\begin{equation}
\boldsymbol{r}_{i,j} = \mathcal{G}_{\theta}\left(\boldsymbol{p}_{i,j}\right),
\end{equation}
where $\mathcal{G}_{\theta}$ is a multi-layer perceptron (MLP) and its parameters $\theta$ are learnable weights. $\boldsymbol{p}_{i,j} = \{\boldsymbol{f}^{l}_{i}, \boldsymbol{f}^{l}_{j}\}$ is a pair of two local appearance patches ($\boldsymbol{f}^{l}_{i}$ and $\boldsymbol{f}^{l}_{j}$) which are $i$-th and $j$-th local appearance patches corresponding to each facial landmark point, respectively.
Each local appearance patches $\boldsymbol{f}^{l}_{i}$ is extracted by the ROI projection which projects a $m\times m$ region around $i$-th landmark point in the input image space to a $m^{'}\times m^{'}$ region on the feature map space.
The same MLP operates on all possible parings of local appearance patches.

\begin{figure}[t]
	\centering
	\includegraphics[width=\textwidth]{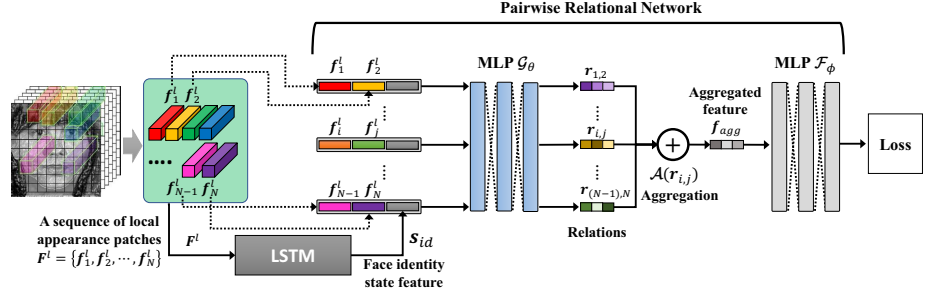}
	\caption{Pairwise Relational Network (PRN). The PRN is a neural network module and takes a set of local appearance patches on the feature maps as input and outputs a single feature vector as its relational feature for the recognition task. The PRN captures unique pairwise relations between pairs of local appearance patches within the same identity and discriminative pairwise relations among different identities}\label{fig:fig_PRN}
\end{figure}

The permutation order of local appearance patches is a critical for the PRN, since without this invariance, the PRN would have to learn to operate on all possible permuted pairs of local appearance patches without explicit knowledge of the permutation invariance structure in the data.
To incorporate this permutation invariance, we constrain the PRN with an aggregation function (Fig. \ref{fig:fig_PRN}):
\begin{equation}
\boldsymbol{f}_{agg} = \mathcal{A}(\boldsymbol{r}_{i,j}) = \sum_{\forall \boldsymbol{r}_{i,j}}{\left(\boldsymbol{r}_{i,j}\right)},
\end{equation}
where $\boldsymbol{f}_{agg}$ is the aggregated relational feature, and $\mathcal{A}(\cdot)$ is the aggregation function which is summation of all pairwise relations among all possible pairing of the local appearance patches.
Finally, a prediction $\widetilde{\boldsymbol{r}}$ of the PRN can be performed with:
\begin{equation}
\widetilde{\boldsymbol{r}}= \mathcal{F}_{\phi}\left(\boldsymbol{f}_{agg}\right),
\end{equation}
where $\mathcal{F}_{\phi}$ is a function with parameters $\phi$, and is implemented by the MLP.
Therefore, the final form of the PRN is a composite function as follows:
\begin{equation}
PRN(\boldsymbol{P}) = \mathcal{F}_{\phi}\left(\mathcal{A}\left(\mathcal{G}_{\theta}\left(\boldsymbol{p}_{i,j}\right)\right)\right),
\label{eq:eq_prn_rev}
\end{equation}
where $\boldsymbol{P} = \{\boldsymbol{p}_{1,2}, \ldots, \boldsymbol{p}_{i,j}, \ldots, \boldsymbol{p}_{(N-1), N}\}$ is a set of all possible pairs of local appearance patches where $N$ denotes the number of local patches on the feature maps.

To capture unique pairwise relations within same identity and discriminative pairwise relations among different identities, a pairwise relation should be identity dependent. So, we modify the PRN such that $\mathcal{G}_{\theta}$ could condition its processing on the identity information. To condition the identity information, we embed a face identity state feature $\boldsymbol{s}_{id}$ as the identity information in the PRN as follows:
\begin{equation}
PRN^{+}(\boldsymbol{P}, \boldsymbol{s}_{id}) = \mathcal{F}_{\phi}\left(\mathcal{A}\left(\mathcal{G}_{\theta}\left(\boldsymbol{p}_{i,j}, \boldsymbol{s}_{id}\right)\right)\right).
\label{eq:equation_prn_idf}
\end{equation}
To get this $\boldsymbol{s}_{id}$, we use the final state of a recurrent neural network composed of LSTM layers and two FC layers that process a sequence of total local appearance patches (\figurename~\ref{fig:fig01_overview}, \ref{fig:fig_idf}).

\subsubsection{Face Identity State Feature}
\begin{figure}[t]
	\centering
	\includegraphics[scale=0.45]{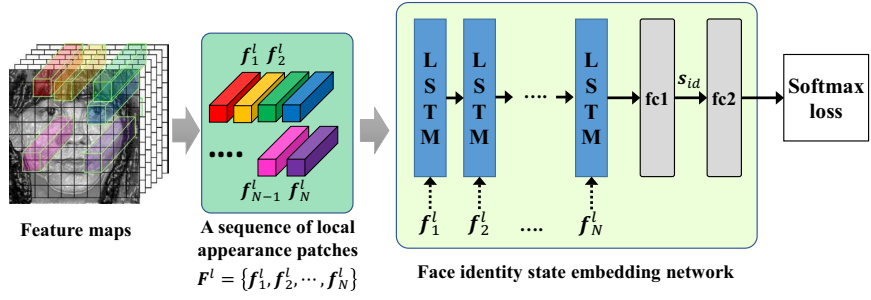}
	\caption{Face identity state feature. A face on the feature maps is divided into $68$ regions by ROI projection around $68$ landmark points. A sequence of local appearance patches in these regions are used to encode the face identity state feature from LSTM networks}
	\label{fig:fig_idf}
\end{figure}

Pairwise relations should be identity dependent to capture unique and discriminative pairwise relations.
Based on the feature maps which are the output of the \textit{conv5\_3} layer in the base CNN model, the face is divided into $68$ local regions by ROI projection around $68$ landmark points. In these local regions, we extract the local appearance patches to model the facial identity state feature $\boldsymbol{s}_{id}$.
Let $\boldsymbol{f}^{l}_{i}$ denote the local appearance patches of $m^{'}\times m^{'}$  $i$-th local region. To encode the facial identity state feature $\boldsymbol{s}_{id}$, an LSTM-based network has been devised on top of a set of local appearance patches $\boldsymbol{F}^{l} = \{\boldsymbol{f}^{l}_{1}, \ldots, \boldsymbol{f}^{l}_{i}, \ldots, \boldsymbol{f}^{l}_{N}\}$ as followings:
\begin{equation}
\boldsymbol{s}_{id} = \mathcal{E}_{\psi}(\boldsymbol{F}^{l}), %FIS_{\omega_{LSTM}}(\boldsymbol{F}^{l}),
\label{eq:equation_lstm}
\end{equation}
where $\mathcal{E}_{\psi}(\cdot)$ is a neural network module which composed of the LSTM layers and two FC layers with learnable parameters $\psi$. We train $\mathcal{E}_{\psi}$ with \textit{softmax} loss function (Fig. \ref{fig:fig_idf}). The detailed configuration of $\mathcal{E}_{\psi}$ used in our proposed method will be presented in Section \ref{sec:implement_details_prn}.

\subsection{Loss Function}
To learn the proposed PRN, we jointly use the triplet ratio loss $L_{t}$, pairwise loss $L_{p}$, and identity preserving loss $L_{id}$ (\textit{softmax}) to minimize distances between faces that have the same identity and to maximize distances between faces that are of different identities:
\begin{equation}
L = \lambda_{1}L_{t} + \lambda_{2}L_{p} + \lambda_{3}L_{id}.
\label{eq:eq_total_loss}
\end{equation}
During training the PRN, we empirically set $\lambda_{1}=1$, $\lambda_{2}=0.5$, and $\lambda_{3}=1$.

\subsubsection{Triplet Ratio Loss}
Triplet ratio loss $L_{t}$ is defined to maximize the ratio of distances between the positive pairs and the negative pairs in the triplets of faces. To maximize $L_{t}$, the Euclidean distances of positive pairs should be minimized and those of negative pairs should be maximized. Let $F(I)\in\mathbb{R}^{d}$, where $I$ is the input facial image, denote the output of a network (the output of $\mathcal{F}_{\phi}$ in the PRN), the $L_{t}$ is defined as follows:
\begin{equation}
L_{t} = \sum_{\forall T}\max\left(0, 1 - \frac{\|F(I_{a}) - F(I_{n})\|_{2}}{\left\|F(I_{a}) - F(I_{p})\right\|_{2} + m}\right),
\label{eq:eq_triplet_loss}
\end{equation}
where $F(I_{a})$ is the output of the network for an anchor face $I_{a}$, $F(I_{p})$ is the output of the network for a positive face image $I_{p}$, and $F(I_{n})$ is the output of the network for a negative face $I_{n}$ in the triplets of faces ${T}$, respectively. $m$ is a margin that defines a minimum ratio in Euclidean space. From recent work by B-.N. Kang \textit{et al.} \cite{PIMNet_CVPRW2017}, they reported that an unbalanced range of distance measured between the pairs of data using only $L_{t}$ during training; this result means that although the ratio of the distances is bounded in a certain range of values, the range of the absolute distances is not. To overcome this problem, they constrained $L_{t}$ by adding the pairwise loss $L_{p}$.

\subsubsection{Pairwise Loss}
Pairwise loss $L_{p}$ is defined to minimize the sum of the squared Euclidean distances between $F(I_{a})$ for the anchor face $I_{a}$ and $F(I_{p})$ for the positive face $I_{p}$. These pairs $I_{a}$ and $I_{p}$ are in the triplets $T$.
\begin{equation}
L_{p} = \sum_{(I_{a}, I_{p}) \in T}\|F(I_{a}) - F(I_{p}) \|_{2}^{2}.
\end{equation}
The joint training with $L_{t}$ and $L_{p}$ minimizes the absolute Euclidean distance between face images of a given pair in the triplets of facs $T$.

\section{Experiments}\label{sec:expt}
The implementation details are given in Section \ref{sec:implement_details}. Then, we investigate the effectiveness of the PRN and the PRN with the face identity state feature in Section \ref{sec:exp_prn}. In Section \ref{sec:exp_lfw}, \ref{sec:exp_ytf}, \ref{sec:exp_ijb-a}, and \ref{sec:exp_ijb-b}, we perform several experiments to verify the effectiveness of the proposed method on the public face benchmark datasets including LFW \cite{LFW}, YTF \cite{YTF_CVPR2011}, IJB-A \cite{IJB-A_CVPR2015}, and IJB-B \cite{IJB-B_CVPRW2017}.

\subsection{Implementation Details}\label{sec:implement_details}
\subsubsection{Training Data}
We used the web-collected face dataset (VGGFace2 \cite{VGG2Face}). All of the faces in the VGGFace2 dataset and their landmark points are detected by the recently proposed face detector \cite{FD_YOON2018} and facial landmark point detector \cite{DAN_CVPRW2017}. We used $68$ landmark points for the face alignment and extraction of local appearance patches. When the detection of faces or facial landmark points is failed, we simply discard the image. Thus, we discarded $24,160$ face images from $6,561$ subjects. After removing these images without landmark points, it roughly goes to $3.1$M images of $8,630$ unique persons. We generated a validation set by selecting randomly about $10\%$ from each subject in refined dataset, and the remains are used as the training set. Therefore, the training set roughly has $2.8$M face images and the validation set has $311,773$ face images, respectively.

\subsubsection{Detailed settings in the PRN}\label{sec:implement_details_prn}
For pairwise relations between facial parts, we first extracted a set of local appearance patches $\boldsymbol{F}^{l} = \{\boldsymbol{f}^{l}_{1}, \ldots, \boldsymbol{f}^{l}_{i}, \ldots, \boldsymbol{f}^{l}_{68}\}$, $\boldsymbol{f}^{l}_{i} \in \mathbb{R}^{1\times1\times2,048}$ from each local region (nearly $1\times1$ size of regions) around $68$ landmark points by ROI projection on the $9\times9\times2,048$ feature maps (\textit{conv5\_3} in \tablename~\ref{tab:tab_baseline_cnn}) in the backbone CNN model. Using this $\boldsymbol{F}^{l}$, we make $2,278$ ($={}^{68}C_{2}$) possible pairs of local appearance patches. Then, we used three-layered MLP consisting of $1,000$ units per layer with batch normalization (BN) \cite{BatchNormal} and rectified linear units (ReLU) \cite{ReLU} non-linear activation functions for $\mathcal{G}_{\theta}$, and three-layered MLP consisting of $1,000$ units per layer with BN and ReLU non-linear activation functions for $\mathcal{F}_{\phi}$. To aggregate all of relations from $\mathcal{G}_{\theta}$, we used summation as an aggregation function.
The PRN is jointly optimized by \textit{triplet ratio} loss $L_{T}$, \textit{pairwise} loss $L_{p}$, and \textit{identity preserving} loss $L_{id}$ (\textit{softmax}) over the ground-truth identity labels using stochastic gradient descent (SGD) optimization method with learning rate $0.10$.
We used mini-batch size of $128$ on four NVIDIA Titan X GPUs. During training the PRN, we froze the backbone CNN model to only update weights of the PRN model.

To capture unique and discriminative pairwise relations dependent on identity, the PRN should condition its processing on the face identity state feature $\boldsymbol{s}_{id}$. For $\boldsymbol{s}_{id}$, we use the LSTM-based recurrent network $\mathcal{E}_{\psi}$ over a sequence of the local appearance patches which is a set ordered by landmark points order from $\boldsymbol{F}^{l}$. In other words, there is a sequence of $68$ length per face. In $\mathcal{E}_{\psi}$, it consist of LSTM layers and two-layered MLP. Each of the LSTM layer has $2,048$ memory cells. The MLP consists of $256$ and $8,630$ units per layer, respectively. The cross-entropy loss with \textit{softmax} was used for training the $\mathcal{E}_{\psi}$ (\figurename~\ref{fig:fig_idf}).

\subsubsection{Detailed settings in the model}
We implemented the base CNN and the PRN models using the Keras framework \cite{Keras_LIB} with TensorFlow \cite{Tensorflow} backend. For fair comparison in terms of the effects of each network module, we train three kinds of models (\textbf{model A}, \textbf{model B}, and \textbf{model C}) under the supervision of cross-entropy loss with \textit{softmax}:
\begin{itemize}
	\item \textbf{model A} is the baseline model which is the base CNN (\tablename{~\ref{tab:tab_baseline_cnn}}).
	\item \textbf{model B} combining two different networks, one of which is the base CNN model (\textbf{model A}) and the other is the $PRN$ (Eq. \eqref{eq:eq_prn_rev}), concatenates the output feature $\boldsymbol{f}^{g}$ of the GAP layer in \textbf{model A} as the global appearance representation and the output of the MLP $\mathcal{F}_{\phi}$ in the $PRN$ without the face identity state feature $\boldsymbol{s}_{id}$ as the local appearance representation. $\boldsymbol{f}^{g}$ is the feature of size $1\times1\times2,048$ from each face image. The output of the MLP $\mathcal{F}_{\phi}$ in the $PRN$ is the feature of size $1\times1\times1,000$. These two output features are concatenated into a single feature vector with $3,048$ size, then this concatenated feature vector is fed into the FC layer with $1,024$ units.
	\item \textbf{model C} is the combined model with the output of the base CNN model (\textbf{model A}) and the output of the $PRN^{+}$ (Eq. \eqref{eq:equation_prn_idf}) with the face identity state feature $\boldsymbol{s}_{id}$. The output of \textbf{model A} in \textbf{model C} is the same of the output in \textbf{model B}. The size of the output in the $PRN^{+}$ is same as compared with the ${PRN}$, but output values are different.
\end{itemize}
All of convolution layers and FC layers use BN and ReLU as nonlinear activation functions except for LSTM layers in $\mathcal{E}_{\psi}$.

\subsection{Effects of the PRNs}\label{sec:exp_prn}
\begin{figure}[t]
	\centering
	\includegraphics[scale=0.335]{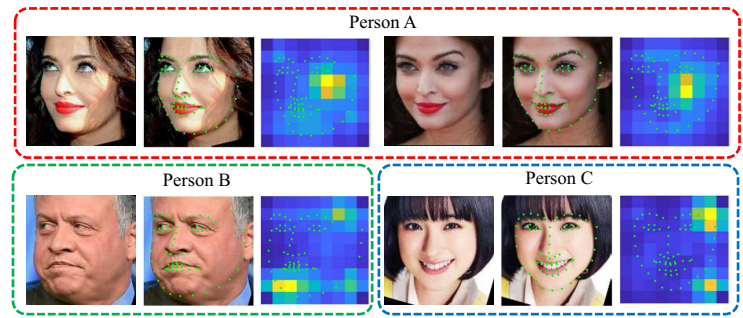}
	\caption{Visualization of the localized facial parts}
	\label{fig:fig_vis_lm_feature}
\end{figure}

To investigate the effectiveness of the PRN and the face identity state feature $\boldsymbol{s}_{id}$, we performed experiments in terms of the accuracy of classification on the validation set during training. For these experiments, we trained two different network models, one of which is a network $PRN$ (Eq. \eqref{eq:eq_prn_rev}) using only the PRN model, and the other is a network $PRN^{+}$ (Eq. \eqref{eq:equation_prn_idf}) using the $PRN$ with the $\boldsymbol{s}_{id}$. We achieved $94.2\%$ and $96.7\%$ accuracies of classification for $PRN$ and $PRN^{+}$, respectively. From these evaluations, when using $PRN^{+}$, we observed that the face identity state feature $\boldsymbol{s}_{id}$ represents the identity property, and the pairwise relations should be dependent on an identity property of a face image. Therefore, these evaluations validate the effectiveness of using the PRN and the importance of the face identity state feature.
We visualize the localized facial parts in \figurename~\ref{fig:fig_vis_lm_feature}, where \textit{Col. 1}, \textit{Col. 2}, and \textit{Col. 3} of each identity are the aligned facial image, detected facial landmark points, and localized facial parts by ROI projection on the feature maps, respectively. We can see that the localized appearance representations are discriminative among different identities.

\subsection{Experiments on the Labeled Faces in the Wild (LFW)}\label{sec:exp_lfw}
We evaluated the proposed method on the LFW dataset, which reveals the \textit{state-of-the-art} of face verification in unconstrained environments.
The LFW dataset is excellent benchmark dataset for face verification in image and contains $13,233$ web crawling images with large variations in illuminations, occlusions, facial poses, and facial expressions, from $5,749$ different identities.
Our models such as \textbf{model A}, \textbf{model B}, and \textbf{model C} are trained on the roughly $2.8$M outside training set (VGGFace2), with no people overlapping with subjects in the LFW. Following the test protocol of \textit{unrestricted with labeled outside data} \cite{LFWTechUpdate}, we test on $6,000$ face pairs by using a squared $L_{2}$ distance threshold to determine classification of \textit{same} and \textit{different}, and report the results in comparison with the \textit{state-of-the-art} methods (\tablename~\ref{tab:lfw_results}).

\begin{table}[t]
	\small
	\centering
	\caption{Comparison of the number of images, the number of networks, the dimensionality of feature, and the accuracy of the proposed method with the \textit{state-of-the-art} methods on the LFW}
	\label{tab:lfw_results}
	\resizebox{\textwidth}{!} {
		\begin{tabular}{lcccc}
			\toprule
			\multicolumn{1}{c}{\textbf{Method}}   							& \textbf{~~Images~~} & \textbf{~~Networks~~} & \textbf{~~Dimension~~} & \textbf{~~Accuracy (\%)~~} \\
			\hline
			%Human                                 							& -                    & -                      & -                          & $97.53$ \\
			DeepFace \cite{DeepFace}              							& $4$M				   & $9$                    & $4,096\times4$             & $97.25$ \\
			DeepID \cite{DeepID}                  							& $202,599$			   & $120$                  & $150\times120$             & $97.45$ \\
			DeepID2+ \cite{DeepID2}               							& $300,000$			   & $25$                   & $150\times120$             & $99.47$ \\
			DeepID3 \cite{DeepID3}                							& $300,000$ 		   & $50$                   & $300\times100$             & $99.52$ \\
			FaceNet \cite{FaceNet}                							& $200$M		   	   & $1$                    & $128$                      & $99.63$ \\
			Learning from Scratch \cite{LFR}      							& $494,414$			   & $2$                    & $160\times2$               & $97.73$ \\
			CenterFace \cite{CenterLoss}         							& $0.7$M			   & $1$                    & $512$                      & $99.28$ \\
			PIMNet${}_{\textrm{TL-Joint~Bayesian}}$ \cite{PIMNet_CVPRW2017} & $198,018$			   & $4$                    & $1,024$                    & $98.33$ \\
			PIMNet${}_{\textrm{fusion}}$ \cite{PIMNet_CVPRW2017}            & $198,018$			   & $4$                    & $6$                        & $99.08$ \\
			SphereFace \cite{SphereFace}          							& $494,414$			   & $1$                    & $1,024$                    & $99.42$ \\
			ArcFace \cite{ArcFace}                							& $3.1$M			   & $1$                    & $512$                      & $99.78$ \\ \midrule
			\textbf{model A} (baseline, only $\boldsymbol{f}^{g}$) 			& $2.8$M			   & $1$                    & $2,048$                    & $\mathbf{99.6}$ \\
			$\mathbf{PRN}$													& $2.8$M			   & $1$ 					& $1,000$ 					 & $\mathbf{99.61}$ \\
			$\mathbf{PRN}^{+}$												& $2.8$M			   & $1$ 					& $1,000$ 					 & $\mathbf{99.69}$ \\
			\textbf{model B} ($\boldsymbol{f}^{g}$ + $PRN$) 				& $2.8$M			   & $1$                    & $1,024$                    & $\mathbf{99.65}$ \\
			\textbf{model C} ($\boldsymbol{f}^{g} + PRN^{+}$) 				& $2.8$M			   & $1$                    & $1,024$                    & $\mathbf{99.76}$ \\ \bottomrule
		\end{tabular}
	}
\end{table}

From the experimental results (Table \ref{tab:lfw_results}), we have the following observations.
First, $PRN$ itself provides slightly better accuracy than the baseline \textbf{model A} (the base CNN model, just uses $\boldsymbol{f}^{g}$) and $PRN^{+}$ outperforms \textbf{model B} which is jointly combined both $\boldsymbol{f}^{g}$ with $PRN$.
Second, \textbf{model C} (jointly combined $\boldsymbol{f}^{g}$ with $PRN^{+}$) beats the baseline model \textbf{model A} by a significantly margin, improving the accuracy from $99.6\%$ to $99.76\%$. This shows that combination of $\boldsymbol{f}^{g}$ and $PRN^{+}$ can notably increase the discriminative power of deeply learned features, and the effectiveness of the pairwise relations between facial local appearance parts (local appearance patches).
Third, compared to \textbf{model B}, \textbf{model C} achieved better accuracy of verification ($99.65\%$ \textit{vs.} $99.76\%$). This shows the importance of the face identity state feature to capture unique and discriminative pairwise relations in the designed PRN model.
Last, compared to the \textit{state-of-the-art} methods on the LFW, the proposed method \textbf{model C} is among the top-ranked approaches, outperforming most of the existing results (Table \ref{tab:lfw_results}). This shows the importance and advantage of the proposed method.

\subsection{Experiments on the YouTube Face Dataset (YTF)}\label{sec:exp_ytf}
We evaluated the proposed method on the YTF dataset, which reveals the \textit{state-of-the-art} of face verification in unconstrained environments.
The YTF dataset is excellent benchmark dataset for face verification in video and contains $3,425$ videos with large variations in illuminations, facial pose, and facial expressions, from $1,595$ different identities, with an average of $2.15$ videos per person. The length of video clip varies from $48$ to $6,070$ frames and average of $181.3$ frames. We follow the test protocol of \textit{unrestricted with labeled outside data}. We test on $5,000$ video pairs and report the test results in comparison with the \textit{state-of-the-art} methods (Table \ref{tab:YTF_results}).

\begin{table}[t]
	\small
	\centering
	\caption{Comparison of the number of images, the number of networks, the dimensionality of feature, and the accuracy of the proposed method with the \textit{state-of-the-art} methods on the YTF}
	\label{tab:YTF_results}
	\resizebox{\textwidth}{!} {
		\begin{tabular}{lcccc}
			\toprule
			\multicolumn{1}{c}{\textbf{Method}}   & \textbf{~~Images~~}  & \textbf{~~Networks~~}  & \textbf{~~Dimension~~}  & \textbf{~~Accuracy (\%)~~} \\
			\hline
			DeepFace \cite{DeepFace}              & $4$M                 & $9$                    & $4,096\times4$          & $91.4$          \\
			DeepID2+ \cite{DeepID2}               & $300,000$            & $25$                   & $150\times120$          & $93.2$          \\
			FaceNet \cite{FaceNet}                & $200$M               & $1$                    & $128$                   & $95.1$          \\
			Learning from Scratch \cite{LFR}      & $494,414$            & $2$                    & $160\times2$            & $92.2$          \\
			CenterFace \cite{CenterLoss}         & $0.7$M               & $1$                    & $512$                   & $94.9$          \\
			SphereFace \cite{SphereFace}          & $494,414$            & $1$                    & $1,024$                 & $95.0$          \\
			NAN \cite{NAN_CVPR2017}               & $3$M                 & $1$                    & $128$                   & $95.7$          \\ \midrule
			\textbf{model A} (baseline, only $\boldsymbol{f}^{g}$)           & $2.8$M               & $1$                    & $2,048$                 & $\mathbf{95.1}$ \\
			$\mathbf{PRN}$ 						  & $2.8$M			     & $1$ 					  & $1,000$ 			    & $\mathbf{95.3}$ \\
			$\mathbf{PRN}^{+}$ 					  & $2.8$M			     & $1$ 					  & $1,000$ 				& $\mathbf{95.8}$ \\
			\textbf{model B} ($\boldsymbol{f}^{g}$ + $PRN$)                     & $2.8$M               & $1$                    & $1,024$                 & $\mathbf{95.7}$ \\
			\textbf{model C} ($\boldsymbol{f}^{g} + PRN^{+}$)                     & $2.8$M               & $1$                    & $1,024$                 & $\mathbf{96.3}$ \\ \bottomrule
		\end{tabular}
	}
\end{table}

From the experimental results (\tablename{~\ref{tab:YTF_results}}), we have the following observations.
First, $PRN$ itself provides slightly better accuracy than the baseline \textbf{model A} (the base CNN model, just uses $\boldsymbol{f}^{g}$) and $PRN^{+}$ outperforms \textbf{model B} which is jointly combined both $\boldsymbol{f}^{g}$ with $PRN$.
Second, $\textbf{model C}$ (jointly combined $\boldsymbol{f}^{g}$ with $PRN^{+}$) beats the baseline model \textbf{model A} by a significant margin, improving the accuracy from $95.1\%$ to $96.3\%$. This shows that combination of $\boldsymbol{f}^{g}$ and $PRN^{+}$ can notably increase the discriminative power of deeply learned features, and the effectiveness of the pairwise relations between facial local appearance patches.
Third, compared to \textbf{model B}, $\textbf{model C}$ achieved better accuracy of verification ($95.7\%$ \textit{v.s.} $96.3\%$). This shows the importance of the face identity state feature to capture unique pairwise relations in the designed PRN model.
Last, compared to the \textit{state-of-the-art} methods on the YTF, the proposed method \textbf{model C} is the \textit{state-of-the-art} ($96.3\%$), outperforming the existing results (Table \ref{tab:YTF_results}). This shows the importance and advantage of the proposed method.

\subsection{Experiments on the IARPA Janus Benchmark A (IJB-A)}\label{sec:exp_ijb-a}
We evaluated the proposed method on the IJB-A dataset \cite{IJB-A_CVPR2015} which contains face images and videos captured from unconstrained environments. It features full pose variation and wide variations in imaging conditions thus is very challenging. It contains $500$ subjects with $5,397$ images and $2,042$ videos in total, and $11.4$ images and $4.2$ videos per subject on average. We detect the faces using face detector \cite{FD_YOON2018} and landmark points using DAN landmark point detector \cite{DAN_CVPRW2017}, and then align the face image with our face alignment method explained in Section \ref{sec:face_alignment}. In this dataset, each training and testing instance is called a `template', which comprises $1$ to $190$ mixed still images and video frames. IJB-A dataset provides $10$ split evaluations with two protocols (1:1 face verification and 1:N face identification). For face verification, we report the test results by using true accept rate (TAR) \textit{vs.} false accept rate (FAR) (\tablename~\ref{tab:ijb-a_results}). For face identification, we report the results by using the true positive identification (TPIR) \textit{vs.} false positive identification rate (FPIR) and Rank-N (\tablename~\ref{tab:ijb-a_results}). All measurements are based on a squared $L_{2}$ distance threshold.

\begin{table}[t]
	\centering
	\caption{Comparison of performances of the proposed PRN method with the \textit{state-of-the-art} on the IJB-A dataset. For verification, TAR \textsl{vs.} FAR are reported. For identification, TPIR \textsl{vs.} FPIR and the Rank-N accuracies are presented}
	\label{tab:ijb-a_results}
	\resizebox{\textwidth}{!} {
		\begin{tabular}{@{}llllllllll@{}}
			\toprule
			\multicolumn{1}{c}{\multirow{2}{*}{Method}}            & \multicolumn{3}{c}{1:1 Verification TAR} & \multirow{2}{*}{} & \multicolumn{5}{c}{1:N Identification TPIR}  \\ \cmidrule(lr){2-4} \cmidrule(l){6-10}
			\multicolumn{1}{c}{}                                   & FAR=0.001        &  FAR=0.01         &  FAR=0.1          &  &  FPIR=0.01        &  FPIR=0.1         &  Rank-1           & Rank-5           &  Rank-10           \\ \midrule
			B-CNN \cite{B-CNN_WACV2016}                            & -                &  -                &  -                &  &  $0.143\pm0.027$  &  $0.341\pm0.032$  &  $0.588\pm0.020$  &  $0.796\pm0.017$  &  -                \\
			LSFS \cite{LSFS_PAMI2017}                              & $0.514\pm0.060$  &  $0.733\pm0.034$  &  $0.895\pm0.013$  &  &  $0.383\pm0.063$  &  $0.613\pm0.032$  &  $0.820\pm0.024$  &  $0.929\pm0.013$  &  -                \\
			DCNN${}_{manual}$+metric \cite{DCNN_Metric_ICCVW2015}  & -                &  $0.787\pm0.043$  &  $0.947\pm0.011$  &  &  -                &  -                &  $0.852\pm0.018$  &  $0.937\pm0.010$  &  $0.954\pm0.007$  \\
			Triplet Similarity \cite{TripletSimilarity_BTAS2016}   & $0.590\pm0.050$  &  $0.790\pm0.030$  &  $0.945\pm0.002$  &  &  $0.556\pm0.065$  &  $0.754\pm0.014$  &  $0.880\pm0.015$  &  $0.95\pm0.007$   &  $0.974\pm0.005$  \\
			Pose-Aware Models \cite{POSE_AWARE_CVPR2016}           & $0.652\pm0.037$  &  $0.826\pm0.018$  &  -                &  &  -                &  -                &  $0.840\pm0.012$  &  $0.925\pm0.008$  &  $0.946\pm0.005$  \\
			Deep Multi-Pose \cite{Deep_Multi_Pose_WACV2016}        & -                &  $0.876$          &  $0.954$          &  &  $0.52$           &  $0.75$           &  $0.846$          &  $0.927$          &  $0.947$          \\
			DCNN${}_{fusion}$ \cite{DCNN_Fusion_WACV2016}          & -                &  $0.838\pm0.042$  &  $0.967\pm0.009$  &  &  $0.577\pm0.094$  &  $0.790\pm0.033$  &  $0.903\pm0.012$  &  $0.965\pm0.008$  &  $0.977\pm0.007$  \\
			Triplet Embedding \cite{TripletSimilarity_BTAS2016}    & $0.813\pm0.02$   &  $0.90\pm0.01$    &  $0.964\pm0.005$  &  &  $0.753\pm0.03$   &  $0.863\pm0.014$  &  $0.932\pm0.01$   &  -                &  $0.977\pm0.005$  \\
			VGG-Face \cite{VGGFACE_BMVC2015}                       & -                &  $0.805\pm0.030$  &  -                &  &  $0.461\pm0.077$  &  $0.670\pm0.031$  &  $0.913\pm0.011$  &  -                &  $0.981\pm0.005$  \\
			Template Adaptation \cite{Template_ADAPT_FGR2017}      & $0.836\pm0.027$  &  $0.939\pm0.013$  &  $0.979\pm0.004$  &  &  $0.774\pm0.049$  &  $0.882\pm0.016$  &  $0.928\pm0.010$  &  $0.977\pm0.004$  &  $0.986\pm0.003$  \\
			NAN \cite{NAN_CVPR2017}                                & $0.881\pm0.011$  &  $0.941\pm0.008$  &  $0.978\pm0.003$  &  &  $0.817\pm0.041$  &  $0.917\pm0.009$  &  $0.958\pm0.005$  &  $0.980\pm0.005$  &  $0.986\pm0.003$  \\
			VGGFace2 \cite{VGG2Face}                               & $0.921\pm0.014$  &  $0.968\pm0.006$  &  $0.990\pm0.002$  &  &  $0.883\pm0.038$  &  $0.946\pm0.004$  &  $0.982\pm0.004$  &  $0.993\pm0.002$  &  $0.994\pm0.001$  \\ \midrule
			\textbf{model A} (baseline, only $\boldsymbol{f}^{g}$) & $\mathbf{0.895\pm0.015}$  &  $\mathbf{0.949\pm0.008}$  &  $\mathbf{0.980\pm0.005}$  &  &  $\mathbf{0.843\pm0.035}$  &  $\mathbf{0.923\pm0.005}$  &  $\mathbf{0.975\pm0.005}$  &  $\mathbf{0.992\pm0.004}$  &  $\mathbf{0.993\pm0.001}$  \\
			\textbf{model B} ($\boldsymbol{f}^{g}$ + $PRN$)        & $\mathbf{0.901\pm0.014}$  &  $\mathbf{0.950\pm0.006}$  &  $\mathbf{0.985\pm0.002}$  &  &  $\mathbf{0.861\pm0.038}$  &  $\mathbf{0.931\pm0.004}$  &  $\mathbf{0.976\pm0.003}$  &  $\mathbf{0.992\pm0.003}$  &  $\mathbf{0.994\pm0.003}$  \\
			\textbf{model C} ($\boldsymbol{f}^{g}$ + $PRN^{+}$)    & $\mathbf{0.919\pm0.013}$  &  $\mathbf{0.965\pm0.004}$  &  $\mathbf{0.988\pm0.002}$  &  &  $\mathbf{0.882\pm0.038}$  &  $\mathbf{0.941\pm0.004}$  &  $\mathbf{0.982\pm0.004}$  &  $\mathbf{0.992\pm0.002}$  &  $\mathbf{0.995\pm0.001}$  \\ \bottomrule
		\end{tabular}
	}
\end{table}

From the experimental results (\tablename{~\ref{tab:ijb-a_results}), we have the following observations.
First, compared to \textbf{model A} (base CNN model), \textbf{model C} (jointly combined $\boldsymbol{f}^{g}$ with $PRN^{+}$) achieved a consistently superior accuracy (TAR and TPIR) on both 1:1 face verification and 1:N face identification.
Second, compared to \textbf{model B} (jointly combined $\boldsymbol{f}^{g}$ with $PRN$), \textbf{model C} achieved also a consistently better accuracy (TAR and TPIR) on both 1:1 face verification and 1:N face identification.
Last, more importantly, \textbf{model C} is trained from scratch and achieves comparable results to the \textit{state-of-the-art} (VGGFace2 \cite{VGG2Face}) which is first pre-trained on the MS-Celeb-1M dataset \cite{MS-Celeb-1M}, which contains roughly 10M face images, and then is fine-tuned on the VGGFace2 dataset. It shows that our proposed method can be further improved by training on the MS-Celeb-1M and our training dataset.

\subsection{Experiments on the IARPA Janus Benchmark B (IJB-B)}\label{sec:exp_ijb-b}
We evaluated the proposed method on the IJB-B dataset \cite{IJB-B_CVPRW2017} which contains face images and videos captured from unconstrained environments.
The IJB-B dataset is an extension of the IJB-A, having $1,845$ subjects with $21.8$K still images (including $11,754$ face and $10,044$ non-face) and $55$K frames from $7,011$ videos, an average of $41$ images per subject.
Because images in this dataset are labeled with ground truth bounding boxes, we only detect landmark points using DAN \cite{DAN_CVPRW2017}, and then align face images with our face alignment method.
Unlike the IJB-A, it does not contain any training splits. In particular, we use the 1:1 Baseline Verification protocol and 1:N Mixed Media Identification protocol for the IJB-B. For face verification, we report the test results by using TAR \textit{vs.} FAR (\tablename~\ref{tab:ijb-b_results}). For face identification, we report the results by using TPIR \textit{vs.} FPIR and Rank-N (\tablename~\ref{tab:ijb-b_results}).
We compare our proposed methods with VGGFace2 \cite{VGG2Face} and FacePoseNet (FPN) \cite{FPN_Align_ICCVW2017}. All measurements are based on a squared $L_{2}$ distance threshold.

\begin{table}[t]
	\small
	\centering
	\caption{Comparison of performances of the proposed PRN method with the \textit{state-of-the-art} on the IJB-B dataset. For verification, TAR \textsl{vs.} FAR are reported. For identification, TPIR \textsl{vs.} FPIR and the Rank-N accuracies are presented}
	\label{tab:ijb-b_results}
	\resizebox{\textwidth}{!} {
		\begin{tabular}{@{}lllllllllll@{}}
			\toprule
			\multicolumn{1}{c}{\multirow{2}{*}{Method}} & \multicolumn{4}{c}{1:1 Verification TAR} & \multirow{2}{*}{} & \multicolumn{5}{c}{1:N Identification TPIR}  \\ \cmidrule(lr){2-5} \cmidrule(l){7-11}
			\multicolumn{1}{c}{}                        			& FAR=0.00001  &  FAR=0.0001  &  FAR=0.001  &  FAR=0.01  &  &  FPIR=0.01  &  FPIR=0.1  &  Rank-1  &  Rank-5  &  Rank-10  \\ \midrule
			VGGFace2 \cite{VGG2Face}                    			& $0.671$  & $0.800$  & $0.0.888$  & $0.949$  &  &  $0.746\pm0.018$  &  $0.842\pm0.022$  &  $0.912\pm0.017$  &  $0.949\pm0.010$  &  $0.962\pm0.007$  \\
			VGGFace2\_ft \cite{VGG2Face}                			& $0.705$  & $0.831$  & $0.908$    & $0.956$  &  &  $0.763\pm0.018$  &  $0.865\pm0.018$  &  $0.914\pm0.029$  &  $0.951\pm0.013$  &  $0.961\pm0.010$  \\
			FPN \cite{FPN_Align_ICCVW2017}              			& -  	   & $0.832$  & $0.916$    & $0.965$  &  &  -  				 &  -  				 &  $0.911$  		 &  $0.953$  		 &  $0.975$  		\\ \midrule
			\textbf{model A} (baseline, only $\boldsymbol{f}^{g}$)  & $\mathbf{0.673}$  & $\mathbf{0.812}$  &  $\mathbf{0.892}$  &  $\mathbf{0.953}$  &  &  $\mathbf{0.743\pm0.019}$  &  $\mathbf{0.851\pm0.017}$  &  $\mathbf{0.911\pm0.017}$  &  $\mathbf{0.950\pm0.013}$  &  $\mathbf{0.961\pm0.010}$  \\
			\textbf{model B} ($\boldsymbol{f}^{g}$ + $PRN$)         & $\mathbf{0.692}$  & $\mathbf{0.829}$  &  $\mathbf{0.910}$  &  $\mathbf{0.956}$  &  &  $\mathbf{0.773\pm0.018}$  &  $\mathbf{0.865\pm0.018}$  &  $\mathbf{0.913\pm0.022}$  &  $\mathbf{0.954\pm0.010}$  &  $\mathbf{0.965\pm0.013}$  \\
			\textbf{model C} ($\boldsymbol{f}^{g}$ + $PRN^{+}$)     & $\mathbf{0.721}$  & $\mathbf{0.845}$  &  $\mathbf{0.923}$  &  $\mathbf{0.965}$  &  &  $\mathbf{0.814\pm0.017}$  &  $\mathbf{0.907\pm0.013}$  &  $\mathbf{0.935\pm0.015}$  &  $\mathbf{0.965\pm0.017}$  &  $\mathbf{0.975\pm0.007}$  \\ \bottomrule
		\end{tabular}
	}
\end{table}

From the experimental results, we have the following observations.
First, compared to \textbf{model A} (base CNN model, just uses $\boldsymbol{f}^{g}$), \textbf{model C} (jointly combined $\boldsymbol{f}^{g}$ with $PRN^{+}$ as the local appearance representation) achieved a consistently superior accuracy (TAR and TPIR) on both 1:1 face verification and 1:N face identification.
Second, compared to \textbf{model B} (jointly combined $\boldsymbol{f}^{g}$ with the $PRN$), \textbf{model C} achieved also a consistently better accuracy (TAR and TPIR) on both 1:1 face verification and 1:N face identification.
Last, more importantly, \textbf{model C} achieved consistent improvement of TAR and TPIR on both 1:1 face verification and 1:N face identification, and achieved the \textit{state-of-the-art} results on the IJB-B.

\section{Conclusion}\label{sec:conclusion}
We proposed a novel face recognition method using the pairwise relational network (PRN) which takes local appearance patches around landmark points on the feature maps, and captures unique pairwise relations between a pair of local appearance patches. To capture unique and discriminative relations for face recognition, pairwise relations should be identity dependent. Therefore, the PRN conditioned its processing on the face identity state feature embedded by the LSTM based network using a sequential local appearance patches. To further improve accuracy of face recognition, we combined the global appearance representation with the PRN. Experiments verified the effectiveness and importance of our proposed PRN and the face identity state feature, which achieved $99.76\%$ accuracy on the LFW, the \textit{state-of-the-art} accuracy ($96.3\%$) on the YTF, comparable results to the \textit{state-of-the-art} for both face verification and identification tasks on the IJB-A, and the \textit{state-of-the-art} results on the IJB-B.

\section*{Acknowledgment}
This research was supported by the MSIT, Korea, under the SW Starlab support program (IITP-2017-0-00897), and ``ICT Consilience Creative program'' (IITP-2018-2011-1-00783) supervised by the IITP.

%\clearpage
\bibliographystyle{splncs04}
\bibliography{egbib}
\end{document}